\pdfoutput=1

\documentclass[11pt]{article}

\usepackage[]{EACL2023}

\usepackage{times}
\usepackage{latexsym}

\usepackage{todonotes}
\usepackage{url}
\usepackage{amssymb}

\def\SPSB#1#2{\rlap{\textsuperscript{{#1}}}\textsubscript{#2}}

\usepackage{graphicx,multirow}
\usepackage{makecell}

\usepackage[T1]{fontenc}

\usepackage[utf8]{inputenc}

\usepackage{microtype}

\usepackage{inconsolata}

\title{Reading and Reasoning over Chart Images for Evidence-based \\ Automated Fact-Checking}

\author{Mubashara Akhtar, Oana Cocarascu \and Elena Simperl \\
        Department of Informatics, King's College London\\
        \texttt{\{mubashara.akhtar,oana.cocarascu,elena.simperl\}@kcl.ac.uk}}

\begin{document}
\maketitle
\begin{abstract}



Evidence data for automated fact-checking (AFC) can be in
multiple modalities such as text, tables, images, audio, or video. 
While there is increasing interest in using images for AFC, previous works mostly focus on detecting manipulated or fake images. 
We propose a novel task, chart-based fact-checking, and introduce ChartBERT as the first model for AFC against chart evidence. ChartBERT leverages textual, structural and visual information of charts to determine the veracity of textual claims. 
For evaluation, we create ChartFC, a new dataset of $15,886$ charts.
We systematically evaluate $75$ different vision-language (VL) baselines and show
that ChartBERT outperforms VL 
models,  
achieving $63.8\%$ accuracy. 
Our results suggest that the task is complex yet feasible, with many challenges ahead.

\end{abstract}

\section{Introduction}



Charts are often used to present data
in news articles, reports, 
scientific publications, and across social media posts \citep{DBLP:journals/cgf/LoGSWBQ22, DBLP:conf/chi/ZhangSPBBP21}. 
For example,
in recent years, charts have been widely used to guide policymakers
in deciding health policies and to communicate COVID information with the general public; a popular example is the coronavirus dashboard by Johns Hopkins University,\footnote{\url{https://coronavirus.jhu.edu/map.html}} which was integrated in several websites \citep{perkel2020behind}.

Misinformation can spread through charts in various ways. 
Previous works in data visualization have discussed how misleading chart design can cause misinformation \citep{DBLP:journals/cgf/LoGSWBQ22}. 
However, a more subtle form of misinformation occurs during chart interpretation 
(e.g. through invalid comparisons, framing correlation as causation, or spreading of misleading claims).
To identify these misinformation types not only the stand-alone chart but the chart together with its message need to be considered jointly \citep{DBLP:journals/cgf/LoGSWBQ22}. 
In this work, we focus on verifying whether charts support or refute claims about them.

\begin{figure}
    \begin{tabular}{ c c }
        \fbox{\begin{minipage}{18.6em}
                \small    
                \textbf{Claim:} Both Thane Baker and Nate Cartmell were ranked last.\\
               \rule{\linewidth}{0.05em}
                  
               \raggedright{\textbf{Evidence:}} \\
                    \centering
                    \includegraphics[scale=0.60]{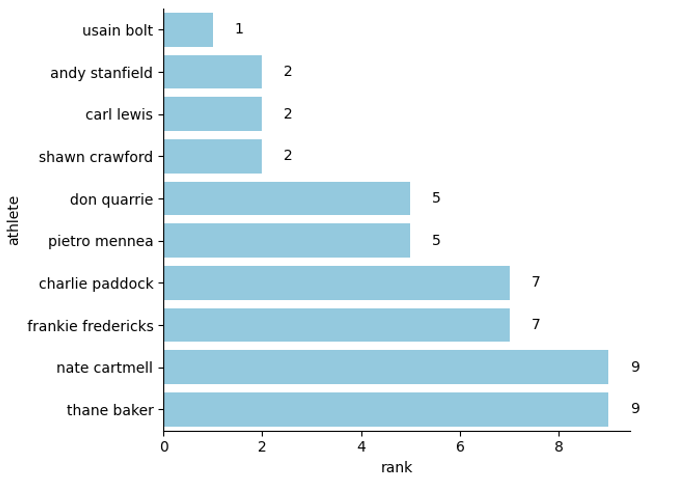}
               \rule{\linewidth}{0.05em}
                \textbf{Label:} Supports\\
        \end{minipage}} & 
    \end{tabular}
    \caption{An example from the ChartFC dataset where the claim is supported by the evidence chart.}
    \label{fig:chart_sample}
\end{figure}



There has been substantial progress in automated fact-checking (AFC) in recent years, 
with a focus on verifying claims against text \citep{wang-2017-liar, thorne-etal-2018-fever, schuster-etal-2021-get, thorne2021evidence, DBLP:journals/corr/abs-2012-00614}, table \citep{DBLP:journals/corr/Aly2021, DBLP:journals/corr/abs-2012-00614, DBLP:conf/iclr/ChenWCZWLZW20, akhtar-etal-2022-pubhealthtab}, and image \citep{DBLP:journals/corr/abs-2205-12487, zlatkova-etal-2019-fact, DBLP:journals/corr/abs-2205-12617} evidence.
Previous work 
has widely ignored claim verification against chart images. 
There are several challenges related to chart fact-checking
as opposed to 
other evidence modalities: 
the structural information, text in charts, and location of text 
need to be considered jointly for chart understanding. 
Text plays a key role and is used,
for example, 
as bar labels, chart titles, or in legends to explain the use of colors. 
Moreover, verifying claims against charts requires 
different reasoning types, e.g. 
retrieving values, finding extremes, or calculating a sum.


To address these challenges, we propose the chart fact-checking task where, given a text claim and a chart, the goal is to classify if it \emph{supports} or \emph{refutes} the claim.
We introduce ChartBERT as the first model for AFC against chart evidence 
comprising $(i)$ an OCR-based reading component to extract text and structural information from chart images; $(ii)$ a sequence generation component to process the extracted information;
and $(iii)$ an encoding component that extends the BERT architecture \citep{devlin-etal-2019-bert} 
with three additional structural embeddings to jointly learn textual and structural representations 
of chart images.   

Moreover, we release Chart{FC} as the first benchmark for chart-based AFC, created using TabFact \cite{DBLP:conf/iclr/ChenWCZWLZW20} as a seed dataset. 
Our dataset contains $15.9k$ human-written claims and 
bars of different colors, orientations, and backgrounds
(see Figure~\ref{fig:chart_sample} for an example). Our highest-performing ChartBERT model achieves $63.8\%$ accuracy on ChartFC.
We compare ChartBERT to $75$ vision-language (VL) baselines, combining five vision encoders, three language encoders, and five fusion methods. 
The best-performing VL model is a transformer-based \citep{DBLP:conf/nips/VaswaniSPUJGKP17}, dual encoder architecture that uses a simple, yet effective fusion block: concatenation and gated recurrent units (GRUs) \citep{DBLP:journals/corr/BahdanauCB14}.
Our results suggest that state-of-the-art 
VL approaches
struggle with the proposed task, calling for 
more research.

Our \textbf{contributions} are as follows: 
$1)$ we propose the chart fact-checking task and build ChartBERT as the first 
chart fact-checking
model; 
$2)$ we introduce Chart{FC}, the first dataset for AFC with chart evidence; $3)$ we systematically evaluate state-of-the-art language/vision encoders and fusion methods on the proposed task, highlighting challenges and providing an analysis of common reasoning types that contribute to failures.\footnote{The ChartFC dataset, trained models, and our code are available at \url{github/link/to/chartfc.com}.} 

\section{Related Work}

\subsection{Verifying Claims against Evidence}
Evidence-based fact-checking aims to predict 
claims' veracity given evidence data. While many datasets focus on text \citep{thorne-etal-2018-fever, kotonya-toni-2020-explainable-automated, schuster-etal-2021-get, wang-2017-liar} and table evidence \citep{DBLP:conf/iclr/ChenWCZWLZW20, gupta-etal-2020-infotabs, DBLP:journals/corr/Aly2021, wang-etal-2021-semeval, akhtar-etal-2022-pubhealthtab}, human fact-checkers use a wider range of modalities for verification \citep{DBLP:conf/ecir/NakovMEBMSAHHBN21, DBLP:journals/corr/abs-2103-12541}. They consult experts and extract information from databases, text, tables, graphics, and audio/video material from 
numerous sources.\footnote{\url{https://ballotpedia.org/The\_methodologies\_of\_fact-checking}} 

Charts 
influence how messages are perceived \citep{DBLP:journals/tvcg/PandeyMNSB14}. 
For example,
\citet{DBLP:conf/chi/LeeYIJS21} 
use the term ``counter-visualization'' to describe 
data visualizations by the anti-vaccination communities in the US
who created charts from publicly available data and interpreted them in a way that challenged the narrative of 
the pandemic, 
leading to
disinformation. 

\subsection{Automated Fact-Checking with Images}

Given that claims and evidence can be conveyed through different modalities, interest in AFC with images has increased recently \citep{nakov2021automated, DBLP:journals/corr/abs-2003-05096, DBLP:journals/corr/abs-2103-12541, DBLP:journals/corr/abs-2205-12487, DBLP:conf/ijcai/SharmaAADMFHSN022}. 
Previous tasks focus mainly on detecting manipulated or fake images 
rather than on evidence-based claim verification 
\citep{blaier-etal-2021-caption, DBLP:conf/nips/KielaFMGSRT20, DBLP:journals/corr/abs-2103-12541, 
DBLP:conf/ijcai/SharmaAADMFHSN022, DBLP:journals/corr/abs-2203-13883}.
Whilst 
manipulated or fake images 
can be detected using the image only, claim verification requires understanding the claim and evidence jointly.

\subsection{Chart Images in Other NLP Tasks}
Two tasks related to chart fact-checking are chart question answering and chart summarization. 
Given a chart image, the summarization task requires to generate a summary of the chart in natural language text \citep{kantharaj-etal-2022-chart, tan2022scientific}.
For question answering (chartQA) the answer to natural language questions is extracted from chart images. 
However, different to claim verification, questions typically provide strong indicators for the correct answers.
Existing chartQA datasets are either small 
\citep{DBLP:conf/chi/KimHA20} or comprise automatically-generated, template-based questions \cite{DBLP:conf/wacv/ChaudhrySGMBJ20, DBLP:conf/iclr/KahouMAKTB18, DBLP:conf/cvpr/KaflePCK18}.

\section{ChartBERT
Model}
\label{sec:chartbert}

\begin{figure}
\centering
\includegraphics[scale=0.47]{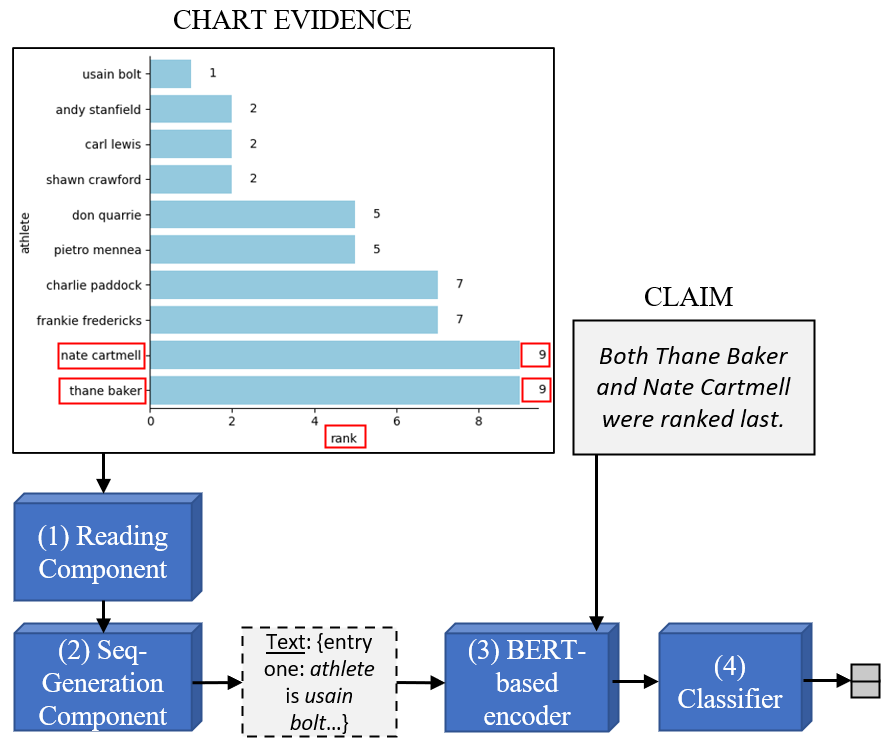}
\caption{\label{fig:chartbert} The ChartBERT architecture.}
\end{figure}

We introduce ChartBERT, a first BERT-based chart fact-checking model.
Our model consists of 
$(i)$ a reading component which extracts text and structural information from charts (Section~\ref{ssec:text_extraction}); 
$(ii)$ a component for generating textual sequences from the information previously extracted (Section~\ref{ssec:sequence_gen}); 
and $(iii)$ a BERT-based encoder with additional structural embeddings for the text extracted from charts  (Section~\ref{ssec:chartbert_encoder}).
The model architecture is shown in Figure~\ref{fig:chartbert}.

\subsection{Task Formulation}

Following previous AFC work \citep{DBLP:conf/iclr/ChenWCZWLZW20, DBLP:journals/corr/Aly2021, thorne-etal-2018-fever, DBLP:conf/semeval/WangMDR21},
we view chart fact-checking as a 
classification task where, given a natural language claim and a piece of evidence (i.e. the chart image), the goal is to decide if the evidence \emph{supports} or \emph{refutes} the claim. 
We use support/refute as labels for claim classification instead of true/false 
as 
we only assess the claim veracity given the provided evidence rather than claiming universal statements.

Each ChartFC sample $i = (c_i, img_i, y_i)$ comprises  a natural language claim $c_i$,  a chart image $img_i$ (see Figure~\ref{fig:chart_sample} for an example), and  a label $y_i \in \{supports, refutes\}$.


\subsection{Reading Text from Chart Images}
\label{ssec:text_extraction}


Given an image $img_i$, the reading component extracts 
text and structural information. 
First, we detect 
text regions in the chart 
using a Tesseract OCR model \citep{kay2007tesseract}.
Specifically, 
for each image, the model extracts 
$n$ text regions $T_i = \{t_1, t_2, ..., t_n\}\SPSB{n}{j=1}$, 
where each region $t_j$ consists of 
$text_j$, a sequence of $m$ tokens, and 
a rectangular bounding box $b_j$ that surrounds the text region in the chart. 
The bounding box is a tuple 
$b_j = (x_j, y_j, w_j, h_j)$
where $x_j$ and $y_j$ are the pixel coordinates of the top left point of the box, and $w_j$ and $h_j$ represent the width and height of the box in 
pixels.
Thus, for each image $img_i$ we obtain the following output $o_i$:  
\[
o_i = f_{R}(img_i) = \{(text_j, x_j, y_j, w_j, h_j)\}\SPSB{n}{j=1} 
\]

\subsection{Text Sequence Generation}
\label{ssec:sequence_gen}

Next, we process the reading component's output into a text sequence $s_i$ consisting of $m$ tokens: 
$$s_i = f_{SeqGen}(o_i) = [s_1, s_2, ... s_m] $$
We
compare two 
approaches 
as follows.

\noindent \textbf{Concatenation:} The concatenation method processes the text regions (i.e. $t_j \in T_i$) based on their coordinates $x_j$ and $y_j$ 
so that texts that are close in the chart are also close 
in the generated sequence. The chart text is concatenated into one sequence and tokens that belong to different text regions are separated using a $[;]$ token. 
Thus, for the chart Figure~\ref{fig:chart_sample} we obtain 
a text sequence starting with ``usain bolt ; 1 ; andy stanfield ; 2 ; [...].''
    
\noindent \textbf{Template:} We use the structural information (i.e. $x$, $y$, $w_j$, $h_j$) to fill templates and generate text sequences. 
We evaluate three templates 
(an example for each template, extracted from Figure~\ref{fig:chart_sample}, is provided in brackets): \\
\noindent $tmp_1$: $\textrm{entry } [num]:[l_x] \textrm{ is } [text_{x}];[l_y] \textrm{ is } [text_{y}]$
(entry one: athlete is usain bolt ; rank is 1); \\
\noindent $tmp_2$: ``$\textrm{row } [num]:[l_x] \textrm{ is } [text_{x}];[l_y] \textrm{ is } [text_{y}]$'' (``row 0: athlete is usain bolt ; rank is 1''); \\
\noindent $tmp_3$: ``$[l_x] \textrm{ is } [text_{x}] \textrm{ when } [l_y] \textrm{ is } [text_{y}]$'' \\ (``athlete is usain bolt when rank is 1'').

The placeholder $[l_x]$ is replaced with the x-axis label from the chart (e.g. ``rank'' in Figure~\ref{fig:chart_sample}). Similarly, the y-axis label (e.g. ``athlete'') replaces $[l_y]$. 
Based on the coordinates, we classify a bounding boxes that contain axes labels (i.e. the boxes with the largest $y$ coordinates).

A counter starting from $one$ replaces $[num]$ and numbers the bars in the chart.
We fill $[text_{y}]$ and and $[text_{x}]$ with text regions detected as bar labels and axis ticks given their positions.  

\begin{figure*}
\centering
\includegraphics[scale=0.6]{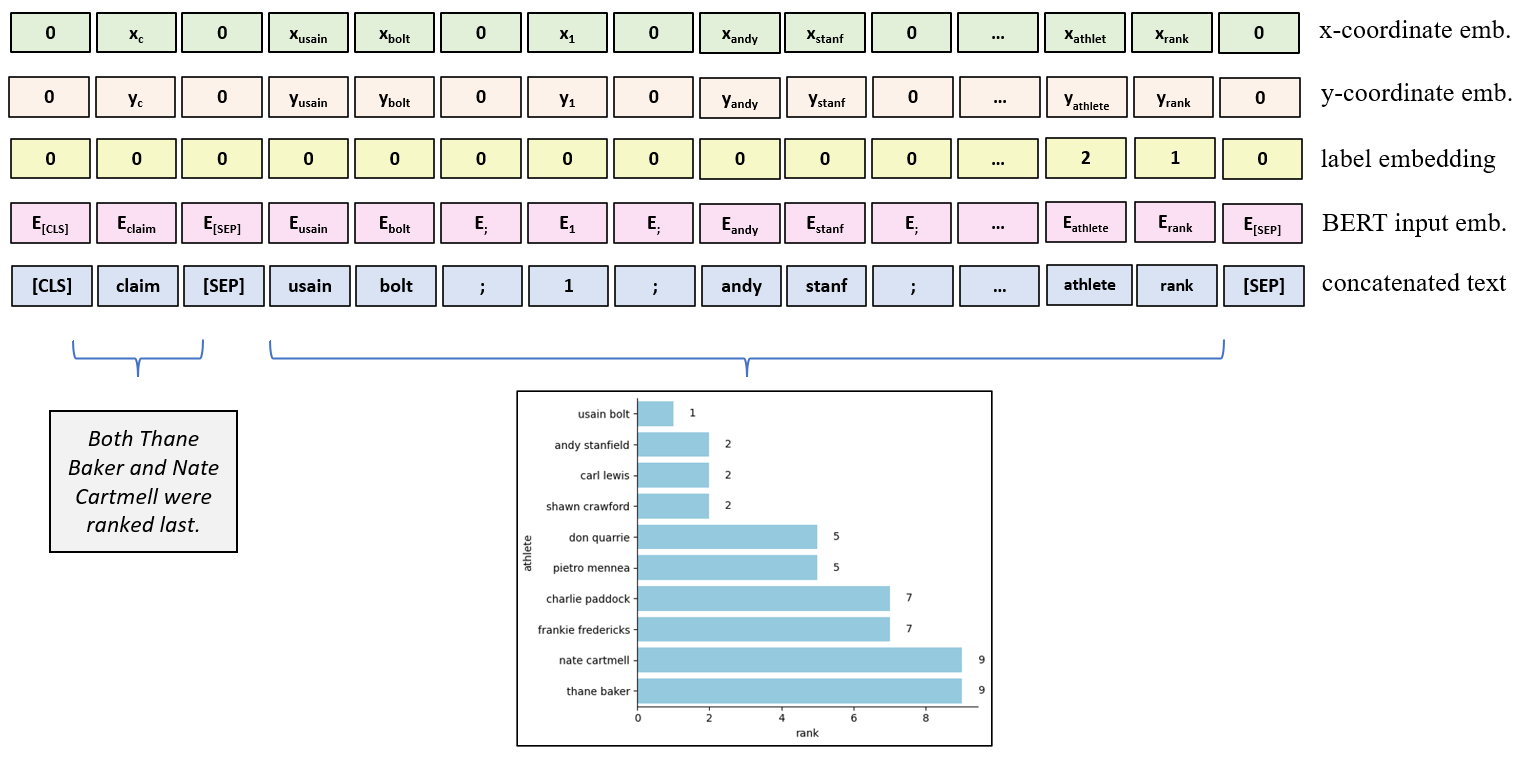}
\caption{\label{fig:bert_encoder} ChartBERT input representation with the text extracted from the chart and concatenated following the approach in Section~\ref{ssec:sequence_gen}. We include additional structural embeddings (i.e. x and y coordinates and label embeddings) to the BERT input embeddings (i.e. token, segment and position embeddings).}
\end{figure*}

\subsection{Encoding and Classification}
\label{ssec:chartbert_encoder}

ChartBERT 
captures the structure of charts through three learned embeddings:
the \emph{x coordinate embedding} which captures the horizontal location of the text in the chart,  
the \emph{y coordinate embedding} which captures the vertical location,  
and the \emph{label embedding} which takes value $1$ if the text region is part of the x-axis label $(l_x)$, $2$ if the text region is part of the for y-axis label $(l_y)$ and $0$ otherwise.

Figure~\ref{fig:bert_encoder} shows an example of the encoder with the structural embeddings.
We concatenate claim $c_i$ and sequence $s_i$, separate them with a $[SEP]$ token, add $[CLS]$ as the first input token, and feed the resulting vector as input to ChartBERT which generates $768$-dimensional representations $h_{i} \in \mathbb{R}\textsubscript{768}$.
Finally, we pass $h_i$ through a fully connected layer and determine the predicted label using sigmoid. ChartBERT uses binary cross entropy to minimize loss on the training set.

$$inp_i = (c_{i}, s_{i}, \{x_j, y_j, l\SPSB{x}{j}, l\SPSB{y}{j}\}\SPSB{n}{j=1})$$
$$ h_i = f_{Encoder}(inp_i)$$
$$ p_i = \sigma(f_{FC}(h_{i})) $$

\section{Evaluation}
\label{ssec:evaluation}


For evaluation, we first create a new dataset, ChartFC.
We compare ChartBERT with several VL baselines, each comprising three components: a vision encoder, a language encoder, and a fusion block to obtain joint representations.
We evaluate the dataset size and potential biases, discuss results obtained with ChartBERT and the baselines, and analyse reasoning types the models fail on.

\subsection{ChartFC Dataset}
\label{ssec:chartfc_dataset}

This section provides an overview of the ChartFC dataset and its creation process.
Each dataset entry comprises a natural language claim, a chart image, and a label $\in \{supports, refutes\}$. 



\subsubsection{The TabFact Dataset}

We use TabFact \citep{DBLP:conf/iclr/ChenWCZWLZW20} as a seed dataset. TabFact is a table fact-checking dataset of natural language claims 
and tables
extracted from Wikipedia as evidence, where the veracity of the claim is decided based on the accompanying table.
Claims were written and evaluated by human crowdworkers 
with at least $95\%$ approval rates for prior tasks and more than $500$ accepted HITs on Amazon Mechanical Turk.
The inter-annotator agreement for the claim verification task is \emph{Fleiss} $\kappa = 0.75$.

\subsubsection{Creation Pipeline}

\begin{figure}
\centering
\includegraphics[scale=0.45]{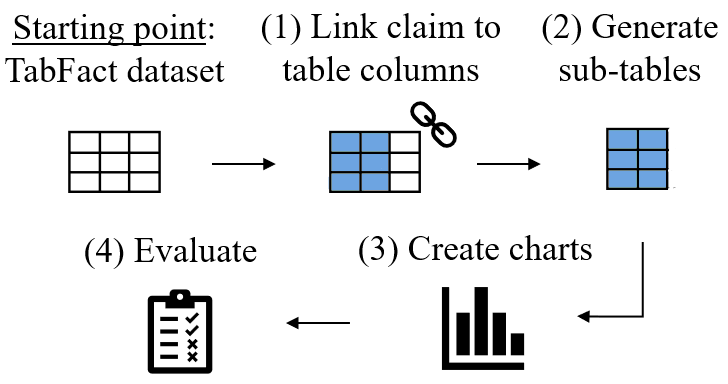}
\caption{\label{fig:dataset_process} Dataset creation process.}
\end{figure}

Figure~\ref{fig:dataset_process} shows the dataset creation process.\footnote{Figure~\ref{fig:dataset_process_example} in the Appendix~\ref{sec:dataset_pipeline_appendix} illustrates the pipeline.} 
Starting with $117,784$ claims and $16,000$ Wikipedia tables from TabFact, we first generate sub-tables. 
To link the claim text to table columns, we $(i)$ lemmatize and tokenize the claim and the table content, 
$(ii)$ link claim tokens to column headers and cells using string matching and heuristic rules, and 
$(iii)$ decide if a claim token is linked to multiple columns using the minimum \emph{Levenshtein distance} \citep{levenshtein1966binary},  and finally, 
$(iv)$  filter sub-tables with a maximum of twenty rows and two linked columns.
This results in a total of $15,886$ pairs of claims and sub-tables. 





Finally, we generate charts using the Python libraries \emph{seaborn} and \emph{matplotlib}.
The charts vary across the dimensions $(i)$ orientation (horizontal, vertical); $(ii)$ bar colors (green, blue, pink); and $(iii)$ background (no/white grid lines, white/gray background color). We show an example in 
Figure~\ref{fig:chart_sample}. 
We partition the dataset into training, validation, and test sets using 8:1:1 ratio and show statistics in Table~\ref{tab:class_distribution}. 

\begin{table}
\centering
\small
\begin{tabular}{lllll}
\hline   & \textbf{Train} & \textbf{Valid} & \textbf{Test} & \textbf{Sum} \\ \hline
Support & 7,048 & 896 & 885 & 8,829  \\
Refute & 5,654 & 697 & 706 & 7,057 \\
\hline
\textbf{Sum}  & 12,702  & 1,593 & 1,591 & 15,886 \\
\end{tabular}
\caption{\label{tab:class_distribution} Class distribution across dataset split.}
\end{table}

\subsubsection{Dataset Evaluation}

To assess the data quality, we apply human and automated evaluation.
We evaluate the sub-table generation step (step $2$ in Figure~\ref{fig:dataset_process}) by checking the verifiability of claims against the extracted sub-tables with TableBERT \citep{DBLP:conf/iclr/ChenWCZWLZW20}.
We obtain $69.3\%$ accuracy on our test set, comparable to $65.1\%$ accuracy reported by \citet{DBLP:conf/iclr/ChenWCZWLZW20} on their test set.

For human validation, we extract $100$ random dataset entries and manually evaluate the claims against sub-tables and charts. Of the $100$ claims, $92$ were successfully verifiable against their sub-tables and chart images, six claims were not verifiable because a relevant column was missing in the sub-table, and 
two claims were already mislabelled in the TabFact dataset. 


\subsubsection{Chart Reasoning Types} \label{subsubsec:charttypes}

\begin{figure}
\centering
\includegraphics[scale=0.65]{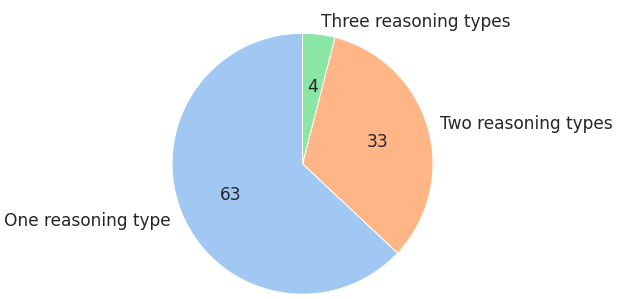}
\caption{\label{fig:reasoning_type_count} Number of chart reasoning types found in $100$ dataset entries.}
\end{figure}

We label $100$ random test samples with \emph{chart reasoning types}, 
using a taxonomy 
of 
common reasoning types humans apply while interacting with data visualisations \citep{DBLP:conf/infovis/AmarES05}. 
We find seven reasoning types present in our data: \emph{retrieve value}, \emph{filter}, \emph{comparison}, \emph{compute derived value}, \emph{find extremum}, \emph{determine range}, and \emph{find anomalies}.\footnote{We describe the chart reasoning types in detail and give examples in Appendix~\ref{sec:appendix_chart_reasoning}.}  
On average, we find $1.4$ different types per claim with most claims including either one or two different reasoning types (see Figure~\ref{fig:reasoning_type_count}). 
The reasoning type \emph{retrieve value}, which requires extracting a value from the chart image given certain criteria, occurs most frequently ($51\%$), followed by \emph{find extremum}, i.e. highest or lowest values in the chart, and \emph{filter}, which occur in approximately a quarter of all labelled claims. More complex types such as \emph{compute derived value} or extracting all values in a given \emph{range} are less frequent. 

\subsection{Vision-Language Baselines}
\label{sec:tv-cfc}

We evaluate our task with several VL baselines, which jointly use claim text and visual information from images for claim verification. 
We also assess the top-$3$ VL baselines with OCR-extracted chart text as additional input. 
Each baseline consists of a language encoder, a vision encoder, and a fusion component to obtain joint representations. 
We systematically evaluate various state-of-the-art encoders and fusion techniques: 
we use shallow (e.g. BERT Embedder \citep{DBLP:conf/eccv/ChenLYK0G0020}) and deep encoders (e.g. DenseNet \citep{DBLP:conf/cvpr/HuangLMW17}), as well as model-agnostic (e.g. concatenation) and model-based (e.g. transformer layers) fusion methods.

\textbf{Language encoders:} 
Given a claim $c_i$, we use a language encoder to obtain a feature vector:
$$
h\SPSB{text}{i} \;\;\; = f_{LangEncoder}(c_i) 
$$
We experiment with three language encoders: \\
    \textbf{BERT Embedder:} Following \citet{DBLP:conf/eccv/ChenLYK0G0020}, we tokenize the claim text into sub-words. For each token, we add the word and position embeddings to obtain the text representation which we then pass through a normalization \citep{ba2016layer} layer. \\
    \textbf{LSTM:} We encode the text with $32$-dimensional word embeddings and pass them through two LSTMs \citep{DBLP:journals/neco/HochreiterS97} with $768$-dimensional hidden states in each layer. We use the hidden states of the second layer as text representations. \\
    \textbf{BERT:} 
    We use a twelve-layer BERT encoder, initialized with weights from a pretrained BERT-base model. 


\textbf{Vision encoders:} We use a vision encoder to extract representations for the chart images:
$$h\SPSB{img}{i} \;\;\; = f_{VisEncoder}(img_i)$$
We evaluate five vision encoders: \\
    \textbf{Fully connected layer:} 
    We use a fully connected layer 
    to extract 768-dimensional representations per image $h\SPSB{img}{i} \;\;\: \in \mathbb{R}\textsubscript{768}$. \\
    \textbf{AlexNet:} Using AlexNet \citep{DBLP:conf/nips/KrizhevskySH12}, for each image, we obtain 
    a representation vector $h\SPSB{img}{i} \;\; \in \mathbb{R}\textsubscript{1024}$ by extracting the model output after the third max pooling layer. \\
    \textbf{ResNet:} We use ResNet-$152$ \citep{DBLP:conf/cvpr/HeZRS16} to obtain $2048$-dimensional image representations 
    by extracting the model output before the two final layers of ResNet-$152$, i.e. before the average pooling layer. \\
    \textbf{DenseNet:} We
    use a DenseNet (DN) \citep{DBLP:conf/cvpr/HuangLMW17} comprising three dense blocks, 
    with $6$, $12$, and $24$ layers, respectively. We extract and concatenate the output of the first and third dense block as low- and high-level feature vectors: 
    $h\SPSB{img}{i} \;\;\: = f_{concat}(f_{DN[block1]}(img_i); f_{DN[block3]}(img_i))$. \\
    \textbf{Vision Transformer (ViT):} We split images into sequences of $n$ $16$x$16$ patches before using them as input to a pretrained base-ViT model \citep{DBLP:conf/iclr/DosovitskiyB0WZ21}.\footnote{\url{https://huggingface.co/google/vit-base-patch16-224}} 
    We extract the hidden states from the model's final layer and use them as image representations, resulting in $768$-dimensional vectors for each patch: $h\SPSB{img}{i} \;\;\, = [h \in \mathbb{R}\textsubscript{768}]_n$.


\textbf{Fusion methods:} We then fuse the text and image representations: 
$$h\SPSB{joint}{i} \;\;\;\;\;\: = f_{Fusion}(h\SPSB{img}{i} \;\;\;\, ; h\SPSB{text}{i} \;\;\;)$$
We experiment with five fusion methods: \\
    \textbf{Concatenation and multiplication:} Concatenation and multiplication are 
    common baseline approaches for multimodal fusion \citep{baltruvsaitis2018multimodal}. We reshape the text and image representations and either $(i)$ concatenate both vectors, or $(ii)$ perform element-wise multiplication.  \\
    \textbf{Concatenation with GRUs:} 
    Inspired by \citet{DBLP:conf/wacv/KafleSPCK20}, we concatenate the text and image representations and pass the resulting vector through $m$ $1$x$1$ convolutional layers and two GRUs. 
    The first GRU takes the input in a forward direction, while the second GRU processes the input vector in a backwards direction to incorporate contextual information:
    $$h\SPSB{concat}{i} \;\;\;\;\;\: = f_{conv}(f_{concat}\{h\SPSB{img}{i} \;\;\;\, ; h\SPSB{text}{i} \;\;\;\}) $$
    $$h\SPSB{joint}{i} \;\;\;\; = f_{concat}\{f\textsubscript{\overrightarrow{\rm GRU}}(h\SPSB{concat}{i} \;\;\;\;\;\;); f\textsubscript{\overleftarrow{\rm GRU}}(h\SPSB{concat}{i} \;\;\;\;\;\;)\} $$ \\
    \textbf{Multimodal Compact Bilinear Pooling (MCB):} MCB is an efficient and popular baseline for multimodal fusion \cite{fukui-etal-2016-multimodal}. 
    The text and image representations are each projected to a higher dimensional space using the projection function Count Sketch \citep{DBLP:journals/tcs/CharikarCF04}. The outer product of the projected vectors is then calculated in Fast Fourier Transform space to obtain a joint representation for both modalities and thus reduce the amount of learnable parameters during model training. \\
    \textbf{Transformer layers:} Given the recent popularity of transformer layers used for joining text and visual representations \citep{tan-bansal-2019-lxmert, DBLP:conf/eccv/ChenLYK0G0020, DBLP:conf/cvpr/YangLW0FWZ0L21-tap}, we use a three-layer transformer to get cross-modal embeddings. 

The representation $h\SPSB{joint}{i}\;\;\;\;$ is passed through two fully-connected layers and sigmoid to obtain the classification. We use binary cross entropy loss and stratified sampling in each training batch to minimize the loss on the training set. 







\subsection{Experimental Setup}

We perform hyper-parameter search on the validation set 
and select the best-performing combination from the following values: 
$\{8, 16, 32\}$ for batch size, $\{1e^{-3}, 7e^{-4}, 5e^{-5}, 5e^{-6}, 5e^{-7}\}$ for learning rate, $\{1, ..., 50\}$ for training epochs with early stopping. 
We also experimented with  different learning rates for the language and vision encoders. 
Ultimately, we used one learning rate for the entire VL model as the modality-specific learning rates did not provide any performance gains.\footnote{The hyper-parameters for each VL baseline can be found in our GitHub repo.}

We run all experiments on a single NVIDIA Tesla V100 GPU with $32GB$ RAM.
We measure model performance with prediction accuracy and (macro) $F_1$ on the test dataset.

\subsection{Results \& Discussion}
\label{ssec:results_discussion}

\noindent \textbf{How does ChartBERT perform on the task? How do different approaches for sequence generation influence model performance?}

Table~\ref{tab:chartbert_results} gives an overview of the  results obtained by ChartBERT. The best ChartBERT variant yields $63.8$\% test accuracy and processes chart text into text sequences using the template $tmp_3$. Compared to the concatenation approach, using $tmp_3$ increases the accuracy by $+3.2\%$. 

Interestingly, the choice of template design impacts the model performance only slightly. 
While template $tmp_3$ might seem more ``natural'' to humans, it does not yield much higher performance compared to $tmp_2$. 


\begin{table}[t]
\centering
\small
\begin{tabular}{|c|c|c|c|c|}
\hline  
\textbf{SeqGen} & \textbf{Val Acc} & \textbf{Val $\mathbf{F_1}$} & \textbf{Test Acc} & \textbf{Test $\mathbf{F_1}$} \\
\hline
concat. & 59.2 & 55.1 & 60.6 & 57.0  \\
temp. $tmp_1$  & 62.4 & 59.1 & 63.3 & 61.0 \\
temp. $tmp_2$  & 62.0 & 59.4 & 61.9 & 58.7 \\
temp. $tmp_3$ & 62.1 & 59.7 & \textbf{63.8} & 61.1 \\
\hline
\end{tabular}
\caption{\label{tab:chartbert_results} Results for ChartBERT  with different sequence generation (SeqGen) approaches: \textbf{concat}enation and \textbf{temp}late.}
\end{table}

\noindent \textbf{How do VL baselines perform on ChartFC? How does the selection of encoder or fusion method impact model performance?}

\begin{table}
\centering
\small
\begin{tabular}{|c|c|c|c|c|}
\hline  
 \textbf{V-Encoder} &   \textbf{Fusion} &  \textbf{no OCR} &  \textbf{text concat} \\
\hline
ViT & concat GRU & 59.8 & 60.5 \\
ResNet & mult & \textbf{60.1} & 61.3 \\
ResNet & concat & 59.8 & \textbf{62.7}  \\
\hline
\end{tabular}
\caption{\label{tab:results_ocr} Test accuracy of top-$3$ VL baselines: without (\textbf{no OCR}) chart text and chart \textbf{text concat}enated. All models use BERT as language encoder.
}
\end{table}

In contrast to many state-of-the-art VL approaches that use simple vision encoders and attention-based fusion \citep{DBLP:conf/eccv/ChenLYK0G0020, DBLP:conf/icml/KimSK21, DBLP:conf/nlpcc/XiaHDZJSCBZ21}, the three best-performing VL models on ChartFC use 
BERT as language encoder, ViT or ResNet to obtain image representations, and either concatenation, multiplication, or concatenation with GRUs as a fusion method.
Using only the claim and chart as input (i.e. without the OCR-extracted chart text), the highest test accuracy we obtain is $60.1$\% with the model consisting of BERT, ResNet, and multiplication fusion (see Table~\ref{tab:results_ocr}).

Regarding the language encoder,\footnote{The complete set of results obtained with different encoders and fusion methods can be found in  Tables~\ref{tab:word_emb_results},~\ref{tab:lstm_results}, and~\ref{tab:bert_results} in the Appendix.} models that use BERT perform best, irrespectively of the vision encoder and fusion method:  the best LSTM-based model achieves $56.1\%$ test accuracy and the best model with BERT embedder yields $56.5\%$ accuracy, both lower than the best BERT-based VL model with $60.1\%$ accuracy.
In contrast, we obtain similar accuracy scores across different vision encoder: for example, replacing ResNet in Table~\ref{tab:results_ocr} row two with a fully connected layer reduces the accuracy slightly by $0.6\%$ to $59.7\%$.
The choice of fusion method does not impact performance strongly: while using multiplication mostly outperforms other methods by a small margin, no fusion method stands out across all vision and language encoders.
We also evaluate the chartQA model PReFIL \citep{DBLP:conf/wacv/KafleSPCK20}, which uses LSTM as language encoder, DenseNet for image representations, and concatenation with GRUs for fusion, and obtain on ChartFC 
a low test accuracy of $55.6\%$.





\noindent \textbf{How does OCR-extracted chart text influence performance of VL models?}

In addition to claim text and chart images used in VL baselines, we also include the text extracted from the charts through OCR as input (see Sections Sections~\ref{ssec:text_extraction} and~\ref{ssec:sequence_gen} for details).
Table~\ref{tab:results_ocr} shows that using the concatenated chart text as input improves accuracy compared to the models that do no use the chart text (e.g. from $59.8\%$ to $62.7\%$). The highest accuracy $62.7\%$ is obtained 
with the BERT-ResNet-concatenation baseline.



\noindent  \textbf{Do models fail on particular chart reasoning types?}

\begin{figure}
\centering
\includegraphics[scale=0.89]{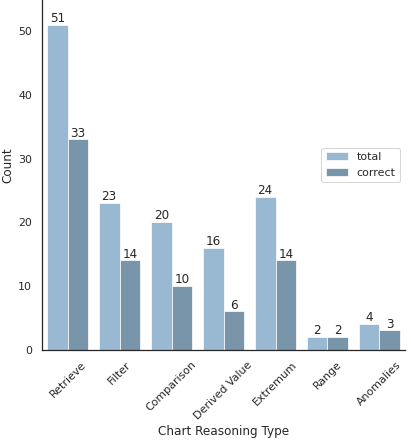}
\caption{\label{fig:chart_reasoning_correct} Chart reasoning types: total count and correct predictions of manually annotated test samples.}
\end{figure}

We evaluate the best VL baseline, consisting of BERT, ViT, and concatenation with GRUs, on the chart reasoning types present in ChartFC and described in Section~\ref{subsubsec:charttypes}. We find that the model performs best on the reasoning types \emph{retrieve value}, \emph{filter}, and \emph{finding extremum}, while struggling particularly with \emph{compute derived values}. 
Figure~\ref{fig:chart_reasoning_correct} shows that the model classifies correctly 65\% (i.e. $33$ out of $51$) of claims that require \emph{retrieval} and 61\% of claims that require \emph{filtering}. 
However, only 50\% of \emph{comparison} claims and $38\%$ of claims required to \emph{compute derived values} are correctly predicted. 
These results 
are in line with previous works that discuss limitations of state-of-the-art models in tasks requiring numerical reasoning capabilities \citep{thawani-etal-2021-representing}.  



\noindent \textbf{Is the dataset size sufficient for our proposed task? Do ChartFC claims contain biases?}

We evaluate the size of the dataset by training our VL baseline (i.e. using BERT, ViT, and concatenation with GRUs) on various subsets of the training data as shown in  Table~\ref{tab:dataset_size_eval} and report the accuracy on the test set. 
The performance on the test set improves as the number of training samples increases. While the performance gain is high when increasing the training set from $1\%$ to $25\%$ (51.6\% accuracy compared to 57\%), the difference in accuracy between the baseline trained on half of the training data and the entire training data is only $2.6\%$, indicating that our training set has a reasonable size.

We also train a claim-only BERT model to determine whether claims contain biases that allow the model to correctly predict the label while ignoring the evidence charts. Trained on the claim text only, the model achieves $52\%$ accuracy on the test set, compared to ChartBERT's accuracy of $(63.8\%)$. We conclude that the claim text itself is not sufficient for correct classification.

\begin{table}
\centering
\small
\begin{tabular}{|c|c|c|}
\hline  
 \textbf{Training Samples} &  \textbf{Test Accuracy} \\  
\hline
 127 (1\%) &  51.6\\
\hline
 3,175 (25\%) & 57.0 \\
\hline
 6,351 (50\%) & 57.1 \\
\hline
 9,526 (75\%) & 58.0 \\
\hline
 12,702 (100\%) & 59.8 \\
\hline
\end{tabular}
\caption{\label{tab:dataset_size_eval} Performance of VL baseline (BERT, ViT, and concatenation with GRUs) with different training set sizes.}
\end{table}

\noindent \textbf{What are the dis-/advantages of an automated dataset pipeline for chart fact-checking?}


We automatically create ChartFC using a table fact-checking dataset as seed  by identifying sub-tables relevant to the claims and then building the charts. ChartFC includes common stylistic variations: bars of different colors, horizontal/vertical orientations, different backgrounds (light/dark, grid lines/no grid lines). While natural charts come with large stylistic variation, using them results in reduced control over task complexity and dataset. In future work, we plan to explore two alternative dataset creation pipelines: first, automated pipelines for other charts types to extend the current dataset, and second, a pipeline with natural charts where we would create claims for charts.

Using 
natural charts would require a multi-step annotation process: 
selecting and separating charts
from other images \citep{DBLP:conf/icwsm/VougiouklisCS20};
writing 
claims which support/refute them; 
evaluating the claims 
to check 
for correctness, typos, 
etc.
We would require annotators with proficiency in interpreting charts, and with basic mathematical and language skills to create claims with different reasoning types (see Figure \ref{fig:reasoning_type_count}).




\section{Conclusion and Future work}

We propose the chart fact-checking task and 
introduce ChartBERT, a novel model for fact-checking claims against chart images comprising three main components: a reading component, a sequence generation component, and an encoder that extends BERT's encoder with structural embeddings. 
We also introduce ChartFC as the first dataset for fact-checking against chart images, consisting of $15,886$ claims and chart images.

ChartBERT achieves 
$63.8\%$ accuracy on ChartFC.
We systematically evaluate $75$ different VL baselines, using various language encoders, vision encoders, and fusion methods. 
The highest-performing VL baseline uses BERT as language encoder, ResNet to extract image representations, and concatenation to obtain joint representations for both modalities. The model achieves $62.7\%$ test accuracy.  
Our results indicate that chart fact-checking, which requires extracting and reasoning over text and structural information from charts, is a challenging task for future research on AFC and VL methods.

\section*{Limitations}

The TabFact dataset \citep{DBLP:conf/iclr/ChenWCZWLZW20} has been a valuable resource for creating ChartFC. However, using it as (the sole) seed dataset 
has limitations.

ChartFC consists of bar charts only; indeed, given the claims and tables found in TabFact, the bar chart was deemed the most appropriate chart type. Various types of charts exist (e.g. pie charts, line charts) and their effectiveness in different data contexts and tasks has been investigated in the literature. For example, \citet{DBLP:journals/tvcg/SaketED19} evaluated the effectiveness of chart types using crowdsourcing experiments across the chart reasoning types we discussed in Section~\ref{subsubsec:charttypes}. In the context of small datasets, i.e. up to 34 rows and two columns which is similar to our setting, \citet{DBLP:journals/tvcg/SaketED19} found bar charts to be the most accurate visualization type for the given chart reasoning types.
In addition to bar charts, other types of charts used as evidence for fact-checking tasks ought to be investigated. \citet{DBLP:journals/cgf/BehrischBKSEFSD18} studied visualization methods for different data types (i.e. multi- and high-dimensional data, relational data, geo-spatial data, sequential and temporal data, and text data). For example, they found that scatter plots were appropriate visualization types for queries regarding data distribution (e.g. correlations and clusters), while line charts were more appropriate for queries about temporal aspects of data. 
To extend ChartFC with other chart types, we require more diverse data types (e.g. sequential and temporal data) and appropriate claims. 



Moreover, ChartFC claims are restricted to English,  whereas misinformation is commonly spread in different languages. 
Future work is necessary to address the limited availability of non-English fact-checking datasets and to contribute to the efforts done in this space \cite{gupta-srikumar-2021-x}.

\bibliography{anthology,custom}
\bibliographystyle{acl_natbib}

\appendix

\section{Dataset Pipeline}
\label{sec:dataset_pipeline_appendix}

\begin{figure*}
\centering
\includegraphics[scale=0.75]{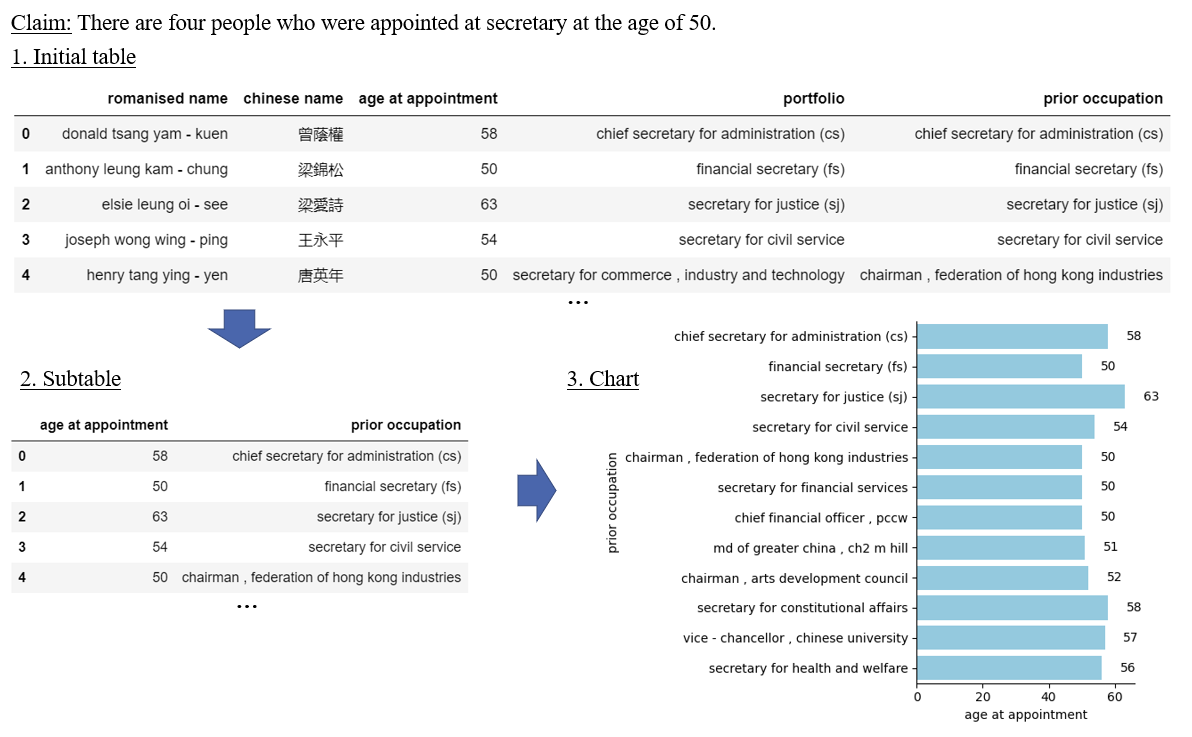}
\caption{\label{fig:dataset_process_example} Example for dataset creation process.}
\end{figure*}

In Figure~\ref{fig:dataset_process_example}, we give an example of the dataset creation pipeline. 
Starting with the claim and initial TabFact table, we first filter columns required to decide the claims veracity label: ``age at appointment'' and ``prior occupation''.
This sub-table is used to create the evidence chart (bottom right). 

\section{Chart Reasoning Types}
\label{sec:appendix_chart_reasoning}

We label $100$ random test set samples with chart reasoning types.
Next, we briefly describe each type, for more details we refer to the taxonomy by \citet{DBLP:conf/infovis/AmarES05}: 

\begin{itemize}
    \item Retrieve Value: Given some conditions, retrieve a single value from the chart image.
    \item Filter: Find all data points in the chart that fulfill some specified conditions. 
    \item Compute Derived Value: Calculate an aggregated value (e.g. average or count) using data points extracted from the chart. 
    \item Find Extremum: Extract the top-$n$ data points given some conditions.
    \item Determine Range: Based on some conditions, find a span of values such that all extracted data points fulfil the conditions.
    \item Find Anomalies: Find any anomalies in a specified set of data points. 
    \item Compare: Compare the values of different data points to each other.
\end{itemize}

\section{VL Baselines}
\label{sec:vl_baselines_appendix}

\begin{table*}[t]
\centering
\small
\begin{tabular}{|c|c|c|c|c|c|c|}
\hline  
\textbf{Lang Encoder} & \textbf{Vis Encoder} & \textbf{Fusion} & \textbf{Val Acc} & \textbf{Val $F_1$} & \textbf{Test Acc} & \textbf{Test $F_1$}\\
 \hline
BERT Emb &	FC & concatenation	 & 56.7 & 37.8 & 55.6 & 36.6 \\
BERT Emb &	FC & concatenation, biGRU	 & 56.2 & 36.0 & 55.6 & 35.7 \\
BERT Emb &	FC & multiplication	 & 56.6 & 52.8 & \textbf{56.5} & 52.3 \\
BERT Emb &	FC & MCB  & 56.2 & 36.1 & 55.6 & 35.7 \\
BERT Emb &	FC & Transformer layers & 56.2 & 36.0 & 55.6 & 35.7 \\
\hline
BERT Emb &	AlexNet & concatenation	 & 56.5 & 40.2 & 55.1 & 38.1 \\
BERT Emb &	AlexNet & concatenation, biGRU	 & 56.2 & 36.0 & 55.6 & 35.7 \\
BERT Emb &	AlexNet & multiplication	 & 57.0 & 41.4 & \textbf{55.9} & 39.9  \\
BERT Emb &	AlexNet & MCB  & 56.2 & 36.0 & 55.6 & 35.7 \\
BERT Emb &	AlexNet	& Transformer layers & 56.2 & 36.0 & 55.6 & 35.7 \\
\hline
BERT Emb &	ResNet 152 & concatenation	  & 56.5 & 45.4 & 56.2 & 45.5 \\
BERT Emb &	ResNet 152 & concatenation, biGRU	& 56.2 & 36.0 & 55.6 & 35.7 \\
BERT Emb &	ResNet 152 & multiplication & 56.6 & 38.3 & \textbf{56.3} & 38.8 \\
BERT Emb &	ResNet 152 & MCB  & 56.2 & 36.0 & 55.6 & 35.7  \\
BERT Emb &	ResNet 152 & Transformer layers & 56.2 & 36.0 & 55.6 & 35.7 \\
\hline
BERT Emb &	DenseNet (6, 12, 24) & concatenation	 & 56.5 & 43.7 & 54.0 & 40.7 \\
BERT Emb &	DenseNet (6, 12, 24) & concatenation, biGRU 	& 56.6 & 45.3 & 54.1 & 42.2 \\
BERT Emb &	DenseNet (6, 12, 24) & multiplication	 & 56.5 & 37.1 & 55.6 & 36.4 \\
BERT Emb &	DenseNet (6, 12, 24) & MCB  & 56.2 & 36.1 & 55.6 & 35.7  \\
BERT Emb &	DenseNet (6, 12, 24) &	Transformer layers  & 56.2 & 36.0 & 55.6 & 35.7  \\
\hline
BERT Emb &	ViT & concatenation & 56.2 & 36.0 & 55.6 & 35.7 \\
BERT Emb &	ViT & concatenation, biGRU	 & 56.2 & 36.0 & 55.6 & 35.7  \\
BERT Emb &	ViT & multiplication	 & 57.1 & 42.1  & 54.8 & 37.6 \\
BERT Emb &	ViT & MCB	& 56.2 & 36.0 & 55.6 & 35.7  \\
BERT Emb &	ViT &	Transformer layers  & 56.2 & 36.0 & 55.6 & 35.7  \\
\hline
\end{tabular}
\caption{\label{tab:word_emb_results} VL baselines using BERT embedder for text encoding, different vision encoders, and fusion methods}
\end{table*}

\begin{table*}
\centering
\small
\begin{tabular}{|c|c|c|c|c|c|c|}
\hline  
\textbf{Lang Encoder} & \textbf{Vis Encoder} & \textbf{Fusion} & \textbf{Val Acc} & \textbf{Val $F_1$} & \textbf{Test Acc} & \textbf{Test $F_1$}\\ \hline
LSTM &	FC 	& concatenation &	56.6 & 36.9 & 55.5 & 35.8 \\
LSTM &	FC & concatenation, biGRU & 56.2 & 36.0 & 55.6 & 35.7 \\
LSTM &	FC & multiplication	& 56.2 & 36.0 & 55.6 & 35.7  \\
LSTM &	FC & MCB	& 56.2 & 36.0 & 55.6 & 35.7 \\
LSTM &	FC & Transformer layers & 56.2 & 36.0 & 55.6 & 35.7 \\
\hline
LSTM &	AlexNet & concatenation & 56.3 & 39.6 & \textbf{56.1} & 39.8 \\
LSTM &	AlexNet & concatenation, biGRU	 & 56.2 & 36.0 & 55.6 & 35.7 \\
LSTM &	AlexNet & multiplication  & 56.2 & 36.0 & 55.6 & 35.7 \\
LSTM &	AlexNet & MCB & 56.2 & 36.0 & 55.6 & 35.7 \\
LSTM &	AlexNet	& Transformer layers & 56.2 & 36.0 & 55.6 & 35.7\\
\hline
LSTM &	ResNet 152 &	concatenation & 56.2 & 36.0 & 55.6 & 35.7 \\
LSTM &	ResNet 152 & concatenation, biGRU & 56.2 & 36.0 & 55.6 & 35.7 \\
LSTM &	ResNet 152 &	multiplication & 56.2 & 36.0 & 55.6 & 35.7  \\
LSTM &	ResNet 152 & MCB & 56.4 & 36.3 & \textbf{56.0} & 35.9 \\
LSTM &	ResNet 152 & Transformer layers & 56.2 & 36.0 & 55.6 & 35.7  \\
\hline
LSTM &	DenseNet (6, 12, 24) & concatenation & 56.2 & 36.0 & 55.6 & 35.7 \\
LSTM &	DenseNet (6, 12, 24) & concatenation, biGRU & 56.2 & 36.0 & 55.6 & 35.7 \\
LSTM &	DenseNet (6, 12, 24) & multiplication	 & 56.2 & 36.0 & 55.6 & 35.7  \\
LSTM &	DenseNet (6, 12, 24) & MCB	 & 56.2 & 36.0 & 55.6 & 35.7 \\
LSTM &	DenseNet (6, 12, 24) &	Transformer layers & 56.2 & 36.0 & 55.6 & 35.7  \\
\hline
LSTM &	ViT & concatenation & 56.2 & 36.0 & 55.6 & 35.7 \\
LSTM &	ViT & concatenation, biGRU	 & 56.2 & 36.0 & 55.6 & 35.7 \\
LSTM &	ViT & multiplication  & 56.2 & 36.0 & 55.6 & 35.7  \\
LSTM &	ViT & MCB & 56.3 & 36.7 & 55.7 & 36.5 \\
LSTM &	ViT &	Transformer layers  & 56.2 & 36.0 & 55.6 & 35.7 \\
\hline
\end{tabular}
\caption{\label{tab:lstm_results} VL baselines with LSTM as language encoder, different vision encoders, and fusion methods}
\end{table*}

\begin{table*}
\centering
\small
\begin{tabular}{|c|c|c|c|c|c|c|}
\hline  
\textbf{Lang Encoder} & \textbf{Vis Encoder} & \textbf{Fusion} & \textbf{Val Acc} & \textbf{Val $F_1$} & \textbf{Test Acc} & \textbf{Test $F_1$}\\ \hline
BERT &	FC & concatenation	 & 59.3 & 50.7 & 59.6 & 51.0 \\
BERT &	FC & concatenation, biGRU	 & 58.8	& 51.1 & 58.5	& 50.2 \\
BERT &	FC & multiplication	 & 59.4 & 54.5 & \textbf{59.7} & 54.9  \\
BERT &	FC & MCB  & 59.7 & 49.6 & 59.1 & 49.3 \\
BERT &	FC & Transformer layers	 & 56.2 & 36.0 & 55.6 & 35.7 \\
\hline
BERT &	AlexNet & concatenation & 	59.5 & 47.9 & 59.1 & 47.6 \\
BERT &	AlexNet & concatenation, biGRU	 & 59.2 & 48.2 & 58.0 &	47.0 \\
BERT &	AlexNet & multiplication	 & 59.0 & 56.2 & \textbf{59.6} & 57.0 \\
BERT &	AlexNet & MCB  & 58.8 & 45.2 & 57.4 & 43.9 \\
BERT &	AlexNet	& Transformer layers & 57.6 & 50.8 &	59.5 & 52.6 \\
\hline
BERT &	ResNet 152 & concatenation	 & 59.8 & 50.9 & 59.8 & 50.8  \\
BERT &	ResNet 152 & concatenation, biGRU	 & 59.1 & 47.0 & 58.8 & 46.7 \\
BERT &	ResNet 152 & multiplication	 & 59.3 & 52.2 & \textbf{60.1} & 53.6 \\
BERT &	ResNet 152 & MCB  & 58.2 & 47.0 & 58.7 & 48.9  \\
BERT &	ResNet 152 & Transformer layers & 56.2 & 36.0 & 55.6 & 35.7 \\
\hline
BERT &	DenseNet (6, 12, 24) &	concatenation 		& 59.1 & 51.4 & \textbf{59.1} & 52.4 \\
BERT &	DenseNet (6, 12, 24) &	concatenation, biGRU &	60.2 & 53.0 & 59.0 & 51.0 \\
BERT &	DenseNet (6, 12, 24) &	multiplication 		& 59.4 & 49.2 & 58.7 & 48.7  \\
BERT &	DenseNet (6, 12, 24) & MCB  & 59.9 & 49.6 & 58.8 & 48.6  \\
BERT &	DenseNet (6, 12, 24) &	Transformer layers & 58.7 &	48.0  & 58.1 & 	46.8 \\
 \hline
BERT &	ViT & concatenation & 56.2 & 36.0 & 55.6 & 35.7\\
BERT &	ViT	 & concatenation, biGRU &	59.0 & 51.2 & \textbf{59.8} & 51.7 \\
BERT &	ViT & multiplication	 & 58.0 & 42.7  & 56.6 & 41.1 \\
BERT &	ViT & MCB	 & 59.2 & 49.5  & 59.2 & 49.6 \\
BERT &	ViT &	Transformer layers & 57.1 & 40.8 & 55.9 & 39.1 \\
\hline 
\end{tabular}
\caption{\label{tab:bert_results} VL baselines with BERT as language encoder, different vision encoders, and fusion methods}
\end{table*}

\begin{figure}
\centering
\includegraphics[scale=0.49]{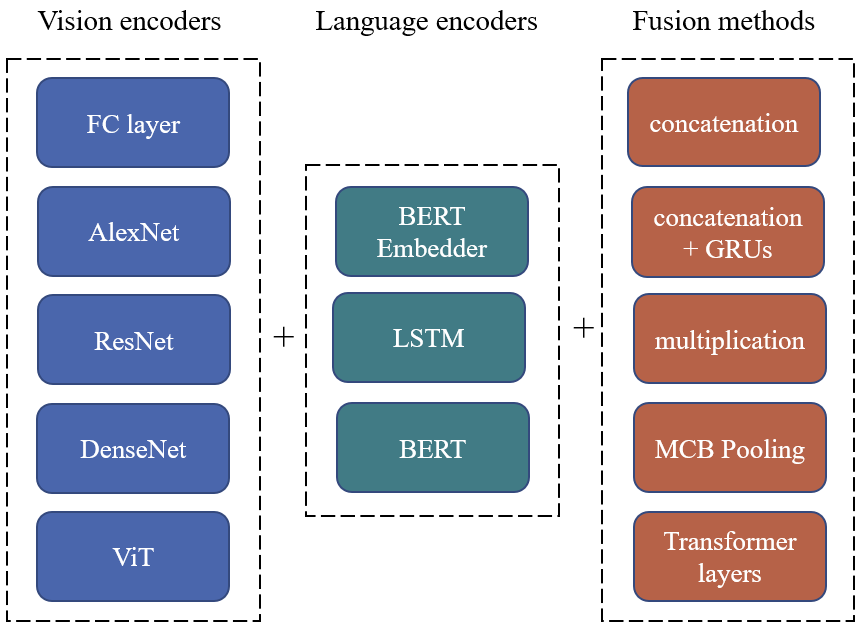}
\caption{\label{fig:vl_components} Encoders and fusion methods used in VL baselines.}
\end{figure}

Figure~\ref{fig:vl_components} provides an overview of all encoders and fusion methods we use in our evaluation. 

Table~\ref{tab:word_emb_results}, \ref{tab:lstm_results}, and \ref{tab:bert_results} provide an overview of all VL baselines we evaluated on ChartFC.



\end{document}